\documentclass[journal]{IEEEtran}
\usepackage{times}
\usepackage{soul}
\usepackage{url}
\usepackage[colorlinks,linkcolor=red,anchorcolor=black,citecolor=black]{hyperref}

\usepackage[utf8]{inputenc}
\usepackage[small]{caption}
\usepackage{graphicx}
\usepackage{subfigure}
\usepackage{amsmath}
\usepackage{amssymb}
\usepackage{amsthm}
\usepackage{mathrsfs}
\usepackage{array}
\usepackage{multirow}
\usepackage{booktabs}
\usepackage{algorithm}
\usepackage{algorithmic}
\usepackage{listings}
\usepackage[frozencache,cachedir=.]{minted}

\usepackage{color}
\definecolor{dkgreen}{RGB}{84,146,181}
\definecolor{gray}{rgb}{0.5,0.5,0.5}
\definecolor{mauve}{rgb}{0.58,0,0.82}
\definecolor{codegreen}{rgb}{0,0.6,0}
\definecolor{codegray}{rgb}{0.5,0.5,0.5}
\definecolor{codepurple}{rgb}{0.58,0,0.82}
\definecolor{backcolour}{rgb}{0.95,0.95,0.92}
\DeclareCaptionFormat{mylst}{\hrule#1#2#3}
\ifCLASSINFOpdf
\else
\fi

\begin{document}
\title{Criteria Comparative Learning for Real-scene Image Super-Resolution}

\author{Yukai Shi\textsuperscript{\dag},
        Hao Li\textsuperscript{\dag},
        Sen Zhang,
        Zhijing Yang,
        Xiao Wang
        % <-this % stops a space

\thanks{
 {\dag} The first two authors share equal contribution.
 }
\thanks{RealSR-Zero dataset: \url{https://github.com/House-Leo/RealSR-Zero}}
\thanks{Code and model: \url{https://github.com/House-Leo/RealSR-CCL}}
}% <-this % stops a space
\maketitle

\begin{abstract}

Real-scene image super-resolution aims to restore real-world low-resolution images into their high-quality versions. A typical RealSR framework usually includes the optimization of multiple criteria which are designed for different image properties, by making the implicit assumption that the ground-truth images can provide a good trade-off between different criteria. However, this assumption could be easily violated in practice due to the inherent contrastive relationship between different image properties. Contrastive learning (CL) provides a promising recipe to relieve this problem by learning discriminative features using the triplet contrastive losses. Though CL has achieved significant success in many computer vision tasks, it is non-trivial to introduce CL to RealSR due to the difficulty in defining valid positive image pairs in this case. Inspired by the observation that the contrastive relationship could also exist between the criteria, in this work, we propose a novel training paradigm for RealSR, named Criteria Comparative Learning (Cria-CL), by developing contrastive losses defined on criteria instead of image patches. In addition, a spatial projector is proposed to obtain a good view for Cria-CL in RealSR. Our experiments demonstrate that compared with the typical weighted regression strategy, our method achieves a significant improvement under similar parameter settings.
\end{abstract}


% \fi
\IEEEpeerreviewmaketitle

\section{Introduction}
\IEEEPARstart{R}{real}-world image super-resolution~\cite{realsr,wang2021unsupervised,sonpami,Lugmayr2020ntire,AIM19} mainly refers to the restoration of real-scene low-resolution images into high-quality ones, which can be obtained by learning a projection function $F(\cdot)$: 

\begin{equation}
I_{o} = F(I_{lf} | \theta),
\end{equation}
where $I_{o}$ and $I_{lf}$ are the output high-resolution and the low-resolution images respectively. $\theta$ is the parameters of $F(\cdot)$. To obtain different resolution images in real-scene configuration,  $I_{hf}$ and $I_{lf}$ are collected by different optical sensors~\cite{xiang2021learning,he2018cascaded,mo2021dense,zhou2020guided,lin2022exploring} with various resolution settings, which is different from the traditional image super-resolution paradigm~\cite{xie2018fast,zhang2020multi,shi2019face,hu2019channel,liu2021cross,shi2017structure} that generates $I_{lf}$ using downsampling techniques. Therefore, compared with the traditional image super-resolution task, RealSR suffers a severer pixel displacement due to the difference between the camera settings to obtain $I_{hf}$ and $I_{lf}$. Although alignment-based methods have been developed to alleviate this problem~\cite{realsr}, current RealSR datasets~\cite{AIM19,Lugmayr2020ntire} still fail to guarantee absolute alignment on pixel-level. 

%% Why this disalignment is the problem for CL? 
%% And here we have not introduce CL -> Maybe we could mention this later
% This disalignment make the contrastive learning algorithm hard to find appropriate positivity / negativity on the $I_{hf}$ and $I_{lf}$ image pair.

%% Why L_wsum could solve the above alignment problem?

In the mainstream RealSR approaches~\cite{kernelgan,shi2020ddet,li2022real}, diverse losses or criteria have been integrated by using their weighted sum to achieve a trade-off between the perceptual- and pixel- similarities:

%% RealSR datasets? 
%% Why therefore?

\begin{equation}
\begin{aligned}
\mathcal{L}_{wsum}={\alpha}\mathbb{C}_{adv}(I_o,I_{hf})& + {\beta} \mathbb{C}_{per}(I_o ,I_{hf}) \\
&+ {\gamma} \mathbb{C}_{pix}(I_o,I_{hf}).
\end{aligned}
\label{eqa:oriloss}
\end{equation}

\begin{figure}[t]
    % \vspace{-2mm}
\centering
\subfigure[Convergence curves of Cria-CL by PSNR index]{
\label{cmt_psnr}
\includegraphics[width=0.9\linewidth]{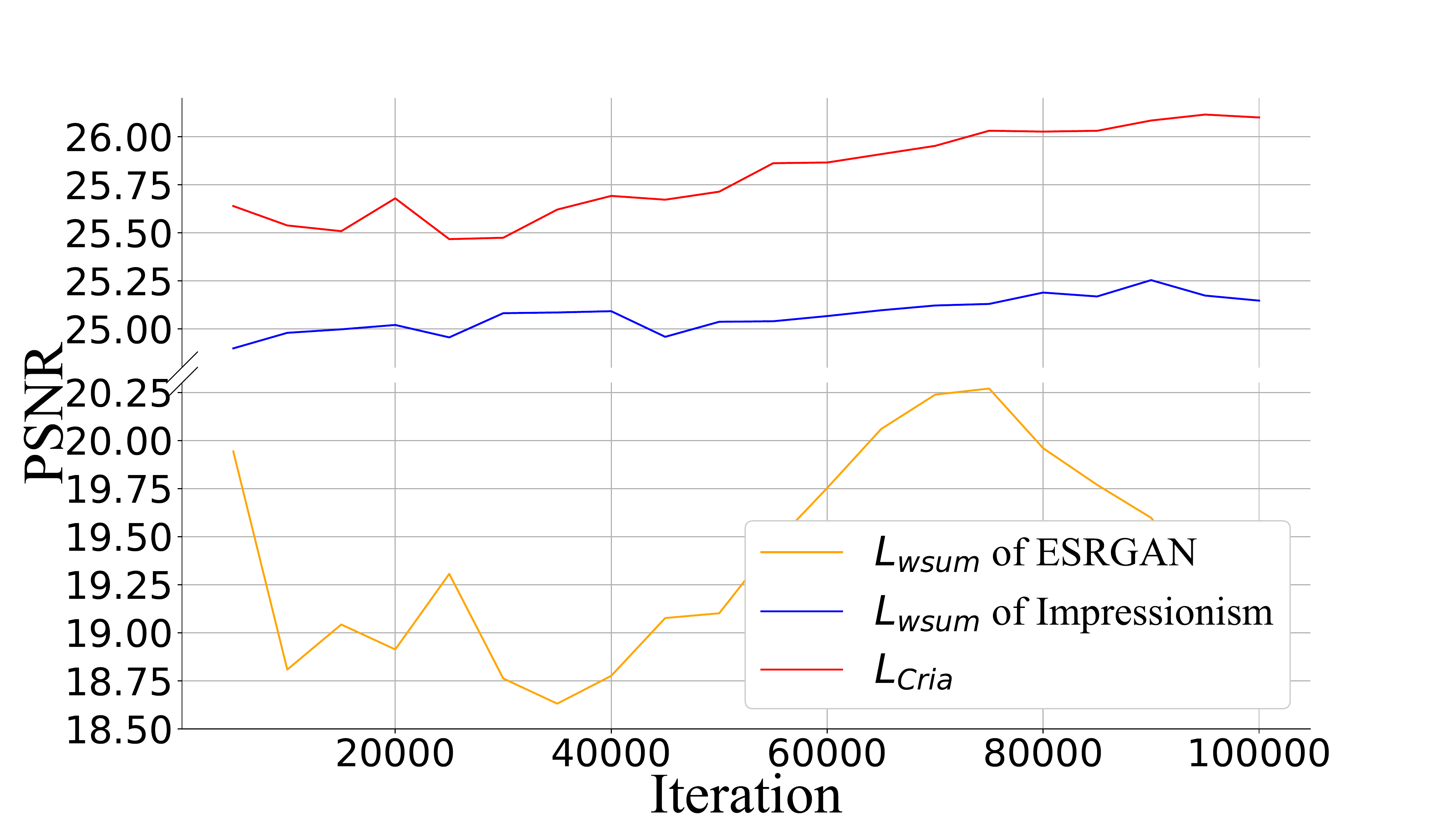}
}
\vspace{-3mm}
\subfigure[Convergence curves of Cria-CL by LPIPS index]{
\label{cmt_lpips}
\includegraphics[width=0.9\linewidth]{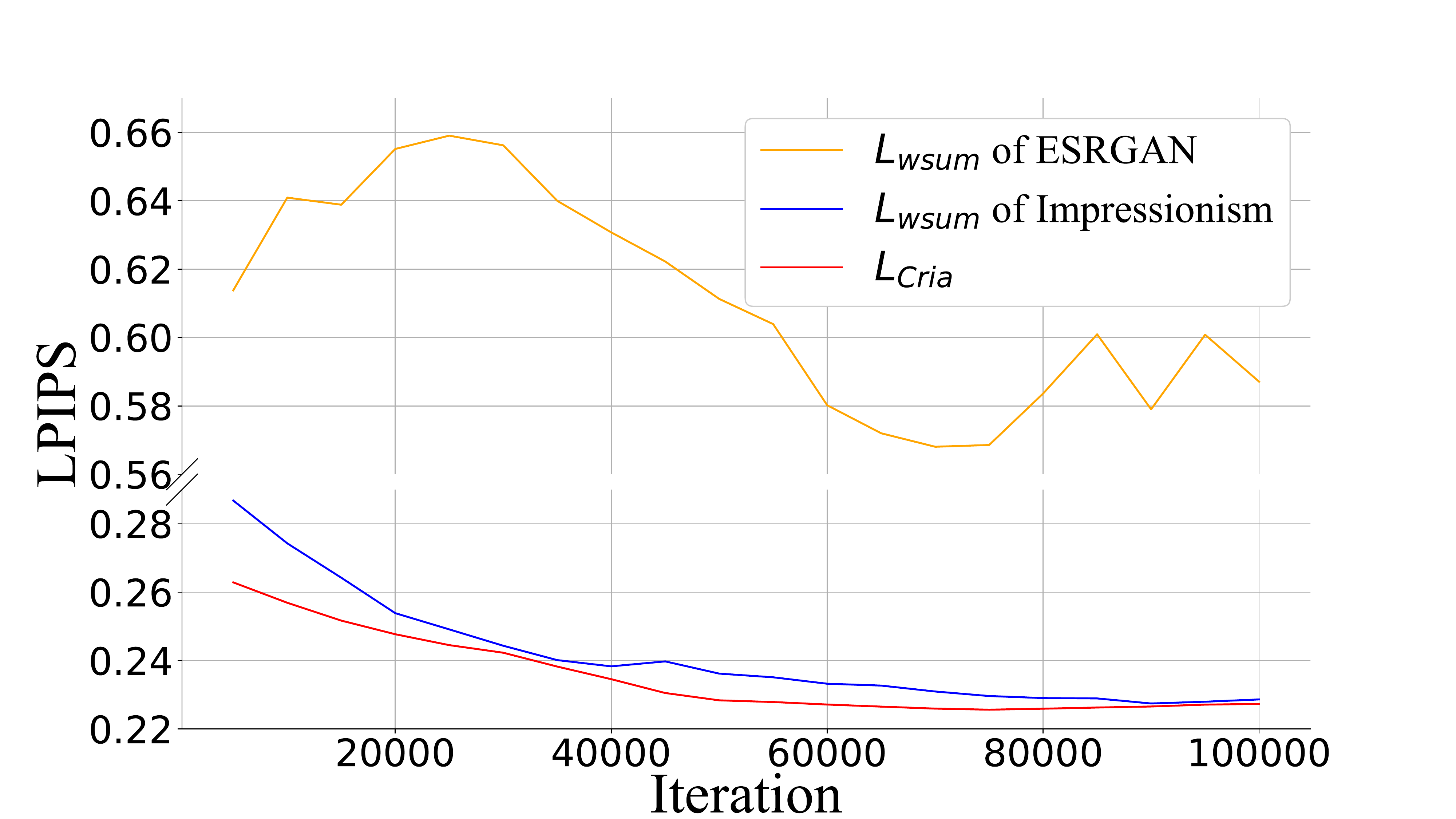}
}
\caption[something short]{Convergence curves of Cria-CL and state-of-the-art methods on NTIRE2020~\cite{Lugmayr2020ntire}. Improved convergence curves are achieved by the criteria comparative learning. }
\label{fig:psnr_lpips}
\vspace{-3mm}
\end{figure}

%% I_o and I_hf, & f and F are not consistent w.r.t. the first paragraph
where $I_o$ is the output of projection function $F$, $[\mathbb{C}_{adv}, \mathbb{C}_{per}, \mathbb{C}_{pix}]$ are the adversarial-, perceptual- and Euclidean- criteria, which focus on restoring different aspects of the images. And $[\alpha, \beta, \gamma]$ are the weights for each loss function, respectively. ESRGAN~\cite{wang2018esrgan} uses ${\mathcal{L}_{wsum}}$ to pursue the trade-off between multiple criteria. SR-VAE~\cite{liu2020photo} employs the KL loss to measure the divergence between latent vectors and standard Gaussian distribution. Similarly, DASR~\cite{wei2020unsupervised} employs generative adversarial nets~\cite{ledig2017photo} to learn the domain distance between the synthetic image and the real-scene image. Then, an Euclidean criterion is used to constrain the bottom image features. These methods implicitly make a strong assumption that the sole ground-truth images can provide a good trade-off between multiple criteria. However, is that a universal solution?

\begin{figure*}[t]
    \centering
    \includegraphics[width=0.9\textwidth]{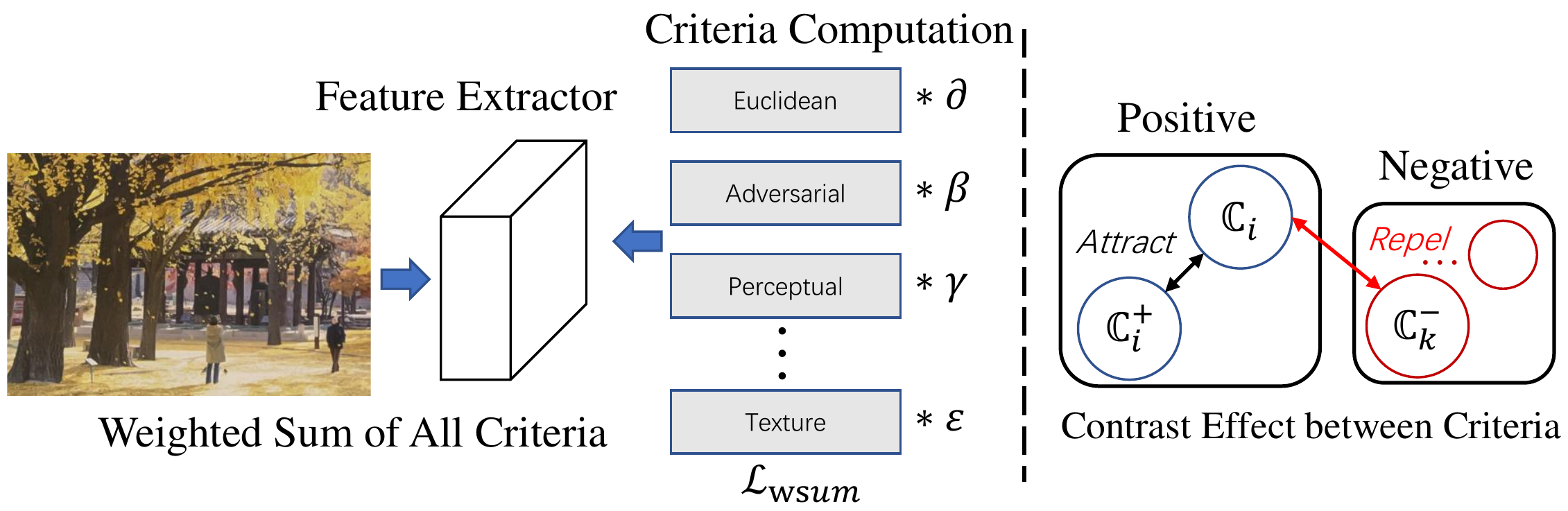}
    \caption{In the mainstream works of RealSR, the multi-task paradigm is widely used by adopting weighted sum of criteria. To arbitrary samples, not all criterion are positive to each other. In this work, we discuss the contrast effect between criteria.}
    \label{fig:cmt}
\end{figure*}

% Will these analyses be added in latter parts?
To answer this question, we re-examine the learning paradigm of typical RealSR pipelines. Accordingly, we found that the ground-truth images are beyond a trade-off between different image properties. For example, suppose we want to generate a realistic and rich texture, the Euclidean criterion plays a positive constraint in the adversarial learning paradigm by regularizing the generative model to preserve a stable and consistent structure. Nevertheless, when it comes to restoring a clear and sharp edge, this generative effect from the adversarial criterion for rich texture plays a \textbf{negative} role against obtaining a sharp edge. In previous works~\cite{wang2018esrgan,yan2021fine}, ${\mathcal{L}_{wsum}}$ is adopted by assuming all criteria are positively contributed to image enhancement. As illustrated in our visual analysis in Fig.~\ref{fig:psnr_lpips} and Fig.~\ref{fig:cmt}, the usage of ${\mathcal{L}_{wsum}}$ tends to achieve a trade-off between the perceptual- and pixel- similarities. Suppose a local region inherently has sharp edge, due to the adversarial criterion takes a considerable proportion, a weighted sum of perceptual- and pixel- criterion often restore a relatively blurry result. This bottleneck motivates us to investigate the contrastive effects among the criteria adaptively.

%% Mention this sentence later when we introduce our method?
% In this paper, we try to explore this contrastive relationship between criteria under the multi-task coexistence condition.  

%% Can some of the this paragraph be put into related work?
The contrastive learning (CL) paradigm~\cite{he2020momentum,khosla2020supervised} provides a promising framework to account for the contrastive relationships, which focus on learning a good feature representation by constructing positive- and negative- training pairs. Specifically, CL attempts to make positive instances close to each other in the high-dimension feature space, while repelling away the negative instances. A basic CL contrastive loss function reads:

\begin{equation}
\mathcal{L}_{CL}=log\frac{\sum_{N}^{i=1}e^{((z_i)^{T} z_i^{+}/\tau)}}{\sum_{K}^{k=1}e^{((z_i)^{T} z_k^{-} /\tau)}},
\label{equ:cons}
\end{equation}
where $z_i$, and $\{z_i^{+}, z_k^{-}\}$ are the hypersphere features of the input anchor sample, and its corresponding positive and negative samples, respectively. $\tau$ is a temperature parameter. Generally, the hypersphere projection of samples is implemented by a convolutional network~\cite{he2020momentum}. In the ImageNet challenge~\cite{deng2009imagenet}, SimCLR~\cite{chen2020simple} obtain the $z_i^{+}$ with data augmentation such as rotation, crop, cutout and resize. And $z_k^{-}$ is an arbitrary sample within the training mini-batch. In image processing tasks like de-raining, SID~\cite{chen2021unpaired} captures the $z_i^{+}$ by searching the clean patch, and the $z_k^{-}$ is a patch which is full of raindrop. 

% Different from the raindrop removal method that style/perceptual level knowledge is enough for adversarial training, 

Although CL has proven successful in many computer vision tasks, however, it remains non-trivial to introduce CL to RealSR, \textbf{due to the difficulty in defining valid positive samples under the RealSR setting}. Specifically, CL methods usually define the positive and negative relationships upon image patches, while in RealSR there are no trivial pixel-level positive samples other than the ground-truth images. Although a ground-truth image can be regarded as perfect positive samples, invalid gradients could occur during optimization when taking the derivative of the \textbf{attached pixel loss: $\left \| I_{hf} - F(I_{lf})\right \|^2_2$.} Moreover, since the ground-truth images have already been used as the labels in Eqn. (2), the repeated use of the ground-truth samples as the input when constructing the contrastive loss could make the network fail to learn the desired discriminative features. Therefore, the positive patches for RealSR are hard to be well defined. 

In this work, we tackle this problem by proposing a novel CL training paradigm for RealSR, named Criteria Comparative Learning (Cria-CL). Inspired by the observation that the inherent contrastive relationship in RealSR also exists between the criteria, e.g., the contrastive effect between the Euclidean- and the adversarial- criterion working on preserving the clear structure and smooth texture simultaneously, Cria-CL attempts to explore such contrastive relationship between criteria by defining the contrastive loss directly on criteria instead of image patches. In addition, in contrast to simply repelling the negative criteria pairs, we formulate the negative contrastive loss using Gaussian potential kernel to introduce uniformity into Cria-CL and provide symmetric context~\cite{cohn2007universally,wang2020understanding}. Furthermore, a spatial projector is developed to obtain a good view for multi-criteria learning to facilitate the training process and enhance the restoration performance.

The contributions are summarized as:

(1).  To explore a new training paradigm for RealSR with appropriate contrastive learning, we build our comparative learning framework upon \textbf{image restoration criteria} (e.g., Euclidean-, perceptual- and adversarial- criterion). 

%the first one has no strong connection with second
(2). In contrast to repelling negative data pair simply, in this paper, we extend the uniformity assumption~\cite{cohn2007universally,wang2020understanding} into criteria to provide fresh symmetric contexts for the multi-task paradigm. %To obtain a good view for feature disentanglement for contrastive multi-task learning, a spatial-view projection paradigm. 

(3). To verify the generalization on out-of-distribution (OOD) data, we built a new RealSR-Zero dataset for evaluation, in which the poor-quality photos are shot by a iPhone4 device and only test image are provided.

(4). Extensive experiments are conducted to verify that each proposed component is solid while the unified framework shows a clear improvement toward state-of-the-art methods.

\section{Related Work}
\textbf{Real-scene Image Super-resolution.}
Different from the traditional image SR that generally focuses on simple synthesized degradation~\cite{wang2018resolution,song2018deeply,zuo2019multi,zhang2021designing}, RealSR~\cite{wei2020component} needs to handle complicated and rough degradation in real-world scenes~\cite{xiang2021learning,lei2020deep}. The first attempt is initially to estimate the degradation process for given LR images, and then apply a paired data-based model for super-resolution. KernelGAN~\cite{kernelgan} proposed to generate blur kernel from label images via a kernel estimation GAN before applying the ZSSR~\cite{zeroshot} method. SSPG~\cite{hu2021image} apply k-nearest neighbors (KNN) matching into neural architecture design. Then, a sample-discriminating learning mechanism based on the statistical descriptions of training samples is used by SSPG to enforce the generative model focus on creating realistic pictures. CDC~\cite{wei2020component} employs a modularized CNN to enhance different cases. SwinIR~\cite{liang2021swinir} investigates a transformer, which gives attractive performance on various image processing tasks. EMASRN~\cite{zhu2021lightweight} facilitates performance with limited parameter number by using an expectation-maximization attention mechanism. TSAN~\cite{zhang2021two} also addresses the attention mechanism in image super-resolution by realizing a coarse-to-fine restoration framework. Wan et al.~\cite{wan2020bringing} applies real-world image restoration model into old photos to build up a practical enhancement framework. Impressionism~\cite{Ji_2020_CVPR_Workshops}, the winner of NTIRE 2020 challenge~\cite{Lugmayr2020ntire}, proposed to estimate blur kernels and extract noise maps from source images and then apply the traditional degradation model to synthesize LR images. Real-ESRGAN~\cite{wang2021realesrgan} introduced a complicated degradation modeling process to better simulate real-world degradation and employ a U-Net discriminator to stabilize the training dynamics. Yet, these methods cannot give out satisfactory results for images with degradations not covered in their model.

To remedy this, several methods try to implicitly grasp the underlying degradation model through learning with the external dataset. DASR~\cite{wei2020unsupervised} proposed a domain-gap aware training strategy to calculate the domain distance between generated LR images and real images that both are used to train the SR model. USR-DA~\cite{wang2021unsupervised} proposed an unpaired SR training framework based on feature distribution alignment and introduced several losses to force the aligned feature to locate around the target domain.

\textbf{Contrastive Learning.} Unsupervised visual representation learning recently achieves attractive success in natural language processing and high-level computer vision tasks~\cite{devlin2018bert,chen2020knowledge,he2020momentum}. Bert~\cite{devlin2018bert} uses masked-LM and next sentence prediction to implement the pre-trained model on a large-scale text dataset. This training strategy contributes to learning general knowledge representations and facilitates reasoning ability in downstream tasks. MoCo~\cite{he2020momentum} revitalizes the self-supervised training for high-level computer vision tasks by proposing momentum contrast learning. Specifically, MoCo builds up positive/negative data queues for contrastive learning, and fills the gap between unsupervised and supervised representation learning.

\textbf{Contrastive Learning for Image Processing.} Many efforts are devoted to contrastive-based image processing tasks. Recently, \cite{park2020contrastive} address the mutual information for various local samples with contrastive learning. \cite{andonian2021contrastive} proposes a novel contrastive feature loss by non-locally patch searching. ~\cite{chen2021unpaired} further explore contrastive feature learning with border tasks by incorporating individual domain knowledge. However, the aforementioned methods still suffer inflexibility of the fixed sample size and a trade-off between different criteria. In this paper, we mainly investigate the feature contrastive learning under multi-task configuration.

\section{Criteria Comparative Learning}

\begin{figure*}[t]
    \centering
    \includegraphics[width=0.97\textwidth]{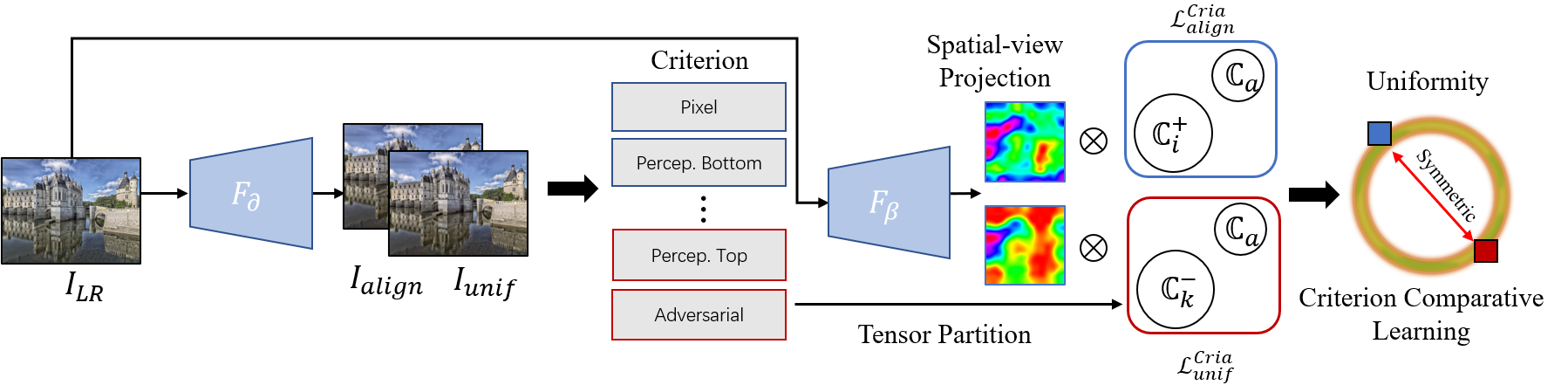}
    \caption{Illustration of criteria comparative learning. Generally, we adopt a multi-task strategy for RealSR. With limited number of criterion, the pixel loss is adopted as the anchor empirically. Each criterion is computed to obtain a tensor individually. We apply a tensor partition to obtain the corresponding negative/positive counterparts, and compute $\mathcal{L}^{Cria}_{unif}$ and $\mathcal{L}^{Cria}_{align}$ with spatial-view masks for optimization. }
    \label{fig:framework}
\end{figure*}

%Since the goal of our work is to improve the perceptual quality of SR images under the real-world setting, we adopt an effective image enhancement backbone for feature extraction, which consists of 23 residual-in-residual dense blocks (RRDBs~\cite{wang2018esrgan}) and incorporate paired-wised training.
%As shown in Fig. , our proposed CMD is consisted of three parts
As the keypoint of this paper is to address the contrast effect between criteria under real-world image setting, we use a simplified RRDB backbone~\cite{wang2018esrgan} for feature extraction and multiple criteria are constructed to emulate a general RealSR framework. As shown in Fig.~\ref{fig:framework}, given a input image $I_{lf}$, we apply a feature extractor $F_{\alpha}(\cdot)$ to produce two intermediate results $\left [ I_{align}, I_{unif} \right ] $ as:

\begin{equation}
\left [ I_{align}, I_{unif} \right ]  = F_{\alpha}(I_{lf} | \theta_{\alpha}).
\label{equ:feature_extraction}
\end{equation}

Typically, in general real-world scene frameworks for image restoration, multiple criteria are adopted as object function for optimization. To realize a criteria comparative learning, we first calculate losses according to each criterion:
\begin{equation}
L_{SR} = \left\{\begin{matrix}
\mathbb{C}_a = L_a(I_{align}, I_{hf})&  \\
\mathbb{C}_i^{+} = L_i^{+}(I_{align}, I_{hf}) &  \\
....  \\
\mathbb{C}_k^{-} = L_k^{-}(I_{unif}, I_{hf}) &  \\
\end{matrix}\right.
\label{eq:loss}
\end{equation}
where $\mathbb{C}_a$ is the anchor criterion. $\mathbb{C}_i^{+}$ and $\mathbb{C}_k^{-}$ are positive and negative criteria toward $\mathbb{C}_a$. Note that the data type of calculated results $\left [ \mathbb{C}_a, .. , \mathbb{C}_i^{+}, .., \mathbb{C}_k^{-} \right ]$  in Equ.~\ref{eq:loss} is tensor, among which we can apply a secondary loss computation and backpropagation. Thus, we can utilize these tensors to realize a criteria comparative learning in RealSR to achieve feature disentanglement. First, we apply Equ.~\ref{equ:cons} into multi-task configuration by replacing the positive/negative patches with \textbf{criteria}:

\begin{equation}
\widetilde{\mathcal{L}}_{Cria}=-log\frac{\sum_{N}^{i=1}e^{(\varphi^{+} (\mathbb{C}_a , \mathbb{C}_i^{+}))}}{ \sum_{K}^{k=1}e^{(\varphi^{-}(\mathbb{C}_a , \mathbb{C}_k^{-}))}},
\label{equ:cons_mtl}
\end{equation}
In addition, $\varphi^{+} $ and $\varphi^{-} $ are similarity measurement function for positive/negative criteria.
 
To enhance interpretability, we further factorize $\widetilde{\mathcal{L}}_{Cria}$ into positive and negative items:

\begin{equation}
\small
\begin{split}
    &\widetilde{\mathcal{L}}_{Cria}=-log\frac{\sum_{N}^{i=1}e^{(\varphi^{+} (\mathbb{C}_a , \mathbb{C}_i^{+}))}}{ \sum_{K}^{k=1}e^{(\varphi^{-}(\mathbb{C}_a , \mathbb{C}_k^{-}))}},\\
    \triangleq &\underbrace {\sum_{N}^{i=1}-\varphi^{+} (\mathbb{C}_a, \mathbb{C}_i^{+} )}_{positive} + \underbrace {log[{\sum_{K}^{k=1}e^{\varphi^{-}(\mathbb{C}_a , \mathbb{C}_k^{-})}} ]}_{negative}. \\
    \triangleq &  \mathcal{L}^{Cria}_{align} + \mathcal{L}^{Cria}_{unif}.
    % L_{all} &= \alpha L_1 + \beta L_{per}+ \gamma L_{adv} + L_{con},
\end{split}
\label{equ:mtl}
\end{equation}
As the $\mathbb{C}$ indicates a computation result of tensors (e.g., $\mathbb{C}_{pix} = ||I_{align} - I_{hf}||_2$), we directly minimizes the loss for positive pair as:

\begin{equation}
\begin{split}
\mathcal{L}^{Cria}_{align}(\mathbb{C}_a,\mathbb{C}^{+};\eta) = \sum_{N}^{i=1}(\mathbb{C}_a(I_{align},I_{hf}) - \mathbb{C}_i^{+}(I_{align},I_{hf}))^{\eta}.
\end{split}
\label{equ:l_align}
\end{equation}
Typical contrastive paradigms simply repel negative data pairs, as shown in Fig.~\ref{fig:framework}, we attempt to realize a uniform distribution on hyperspherical space to provide symmetric context. Instead of repelling negative criteria irregularly, the criteria are enforced to reach a uniform distribution~\cite{wang2020understanding} on the hypersphere. Different from the uniformity assumption in~\cite{wang2020understanding}, we realize $\varphi^{-}$ by proposing the following uniformity loss for negative criteria to provide symmetric context:

\begin{equation}
\begin{aligned}
\mathcal{L}^{Cria}_{unif}(\mathbb{C}_a,\mathbb{C}^{-};t) = -log[\sum_{K}^{k=1}e^{({-t(\mathbb{C}_a(I_{unif},I_{hf})  + \mathbb{C}_{k}^{-}(I_{unif},I_{hf}))^2)} }].
\end{aligned}
\label{eqa:l_uniform}
\end{equation}
  
\textbf{Spatial-view Projection.}
A good viewpoint for criterion disentanglement is non-trivial for contrastive learning. In RealSR, an image contains rich texture, which often lead to $\left [ \mathbb{C}_a , \mathbb{C}_k^{-} \right ]$ not in the same distribution space diametrically. Hence, its unreasonable to take contrastive multi-task learning for the whole image without looking into special cases. Searching a local patch with a fixed size for spatial view has inherent inflexibility. Thus, we apply a non-local spatial-view search and projection:
\begin{equation}
I_{SR} = I_{align} * \mathbb{S} + I_{unif} * \widetilde{\mathbb{S}},
\end{equation}
where $[\mathbb{S},\widetilde{\mathbb{S}}]$ are spatial masks for $I_{unif}$ and $I_{align}$, we obtain them by extracting multi-task oriented feature representations and the original image with $F_{\beta}$. Then, we apply $[\mathbb{S},\widetilde{\mathbb{S}}]$ by the class activation map (CAM)~\cite{zhou2016learning} fashion to realize spatial projection for positive and negative criterion, and present final output $I_{SR}$. As illustrated in Fig.~\ref{fig:framework}, the pairwise criteria are jointly optimized with $\left [F_{\alpha}, F_{\beta}  \right ]$ for a comparative learning.

\textbf{Model Details.}
We apply RRDBs as the backbone~\cite{wang2018esrgan} in feature extractor $F_{\alpha}$. Specifically, we added two sub-branches at the end of RRDBs, each sub-branch consists of three Residual blocks~\cite{residual_net}. Then, we send the intermediate output of RRDBs into two sub-branches, each sub-branch uses different loss (e.g. L1 loss and adversarial loss) for optimization, and produces $I_{align}$ and $I_{unif}$ respectively. In addition, we send the original image $I_{LR}$ into another feature extractor $F_{\beta}$, which consists of three Residual blocks, an upsampling operation and a softmax operation to obtain the spatial masks $[\mathbb{S},\widetilde{\mathbb{S}}]$. The upsampling operation is to keep the spatial masks $[\mathbb{S},\widetilde{\mathbb{S}}]$ with the same spatial sizes ($H \times W$) as $[I_{align},I_{unif}]$. 

\textbf{Anchor Selection.} How to choose a fixed anchor criterion and corresponding negative/positive counterpart is a critical issue in our algorithm. With a limited criterion number, we successively pick up pixel-, adversarial- and perceptual- criterion as anchor to observe the experimental result. As depicted in Tab.~\ref{table:ablation_anchor}, by adopting $\mathbb{C}_{adv}$ as anchor criterion, our model shows a poor results. Since a pure adversarial loss often performs unsteadily during training, which causes all criteria to become positive counterparts. As shown in Tab.\ref{table:ablation_anchor}, once we set any negative criterion for $\mathbb{C}_{adv}$ as the negative item, the performance becomes poor. %Thus, we set $\mathbb{C}_{pix}$ as anchor for positive/negative counterpart partition.
\begin{figure}[t]
\begin{minted}[
fontsize=\fontsize{7}{7.5},
frame=single,
framesep=2.5pt,
baselinestretch=1.05,
]{python}
# x: anchor criterion
# y: positive criterion
# z: the list of all negative criteria
# S, hat_S: Spatial-view in Eq. 8
# lam: empirical factor
def lalign(l_anchor, l_pos_list, alpha=2):
    for l_pos in l_pos_list:
        l_align += (l_anchor - l_pos).pow(alpha)
        return l_align

def lunif(l_anchor, l_neg_list, t=2):
    l_unif=0
    for l_neg in l_neg_list:
        l_unif += (l_anchor + l_neg).pow(t).mul(-t).exp()
    l_unif = -lunif.log()
    return l_unif

loss = S * lalign(x, y) + lam * hat_S * lunif(x, z) / 2
\end{minted}
\caption{Pytorch implementation of $\mathcal{L}^{Cria}_{align}$ and $\mathcal{L}^{Cria}_{unif}$.}
\label{fig:pytorch-losses-code}%
\vspace{-20pt}
\end{figure}
Literally, Euclidean criterion can find distinct positive/negative examples and presents a solid performance. We therefore use Euclidean criterion as $\mathbb{C}_a$ empirically to illustrate our framework. Since the pixel loss is set as the anchor, we use $[\mathbb{C}_{ssim},\mathbb{C}^{B}_{perc}]$) as positive items because they all based on pixel similarity. As $[\mathbb{C}^{T}_{perc},\mathbb{C}_{adv}]$ have potential to produce arbitrary texture/artifact, which often go against to the sharpness of the structure, we use them as negative items. Its note that we have employed a spatial-view projection, thus only regional pixels rather than full image will be handle with the criterion comparative learning. 

To this end, we can realize a criteria partition as:  $\mathcal{L}^{Cria}_{align}(\mathbb{C}_{pix},[\mathbb{C}_{ssim},\mathbb{C}^{B}_{perc}];\eta)$ and $\mathcal{L}^{Cria}_{unif}(\mathbb{C}_{pix}, [\mathbb{C}^{T}_{perc},\mathbb{C}_{adv}];t )$, where $\mathbb{C}^{B}_{perc}$ and $\mathbb{C}^{T}_{perc}$ are bottom- and top- feature index of VGG-19. And $\left [ \eta, t \right ]$ are used to determine the loss landscapes, we follow prior works~\cite{wang2020understanding} to set those two values empirically.
To the $\mathbb{C}^{T}_{perc}$, we assume the perceptual constraint toward realistic style needs to be disentangled from rough pixel similarity.

Follow the prior work~\cite{wang2018esrgan}, the overall loss function consists of pixel loss $\mathcal{L}_{pix}$, perceptual loss $\mathcal{L}_{per}$, adversarial loss $\mathcal{L}_{adv}$, $\mathcal{L}^{Cria}_{align}$ and $\mathcal{L}^{Cria}_{unif}$, which can be expressed as follows:
\begin{equation}
% \small
\begin{aligned}
    \mathcal{L}(I_{SR},I_{GT})&=\alpha \mathcal{L}_{pix}(I_{SR},I_{GT}) + \beta \mathcal{L}_{per}(I_{SR} ,I_{GT}) \\
&+ \gamma \mathcal{L}_{adv}(I_{SR},I_{GT}) + \lambda_{a} \mathcal{L}^{Cria}_{align} + \lambda_{u} \mathcal{L}^{Cria}_{unif}.
\end{aligned}
\end{equation}

We set the $\alpha=0.01$, $\beta=1$, $\gamma=0.005$, $\lambda_{a}=\lambda_{u}=0.01$.

\begin{figure*}[t]
    \centering
    \includegraphics[width=0.92\textwidth]{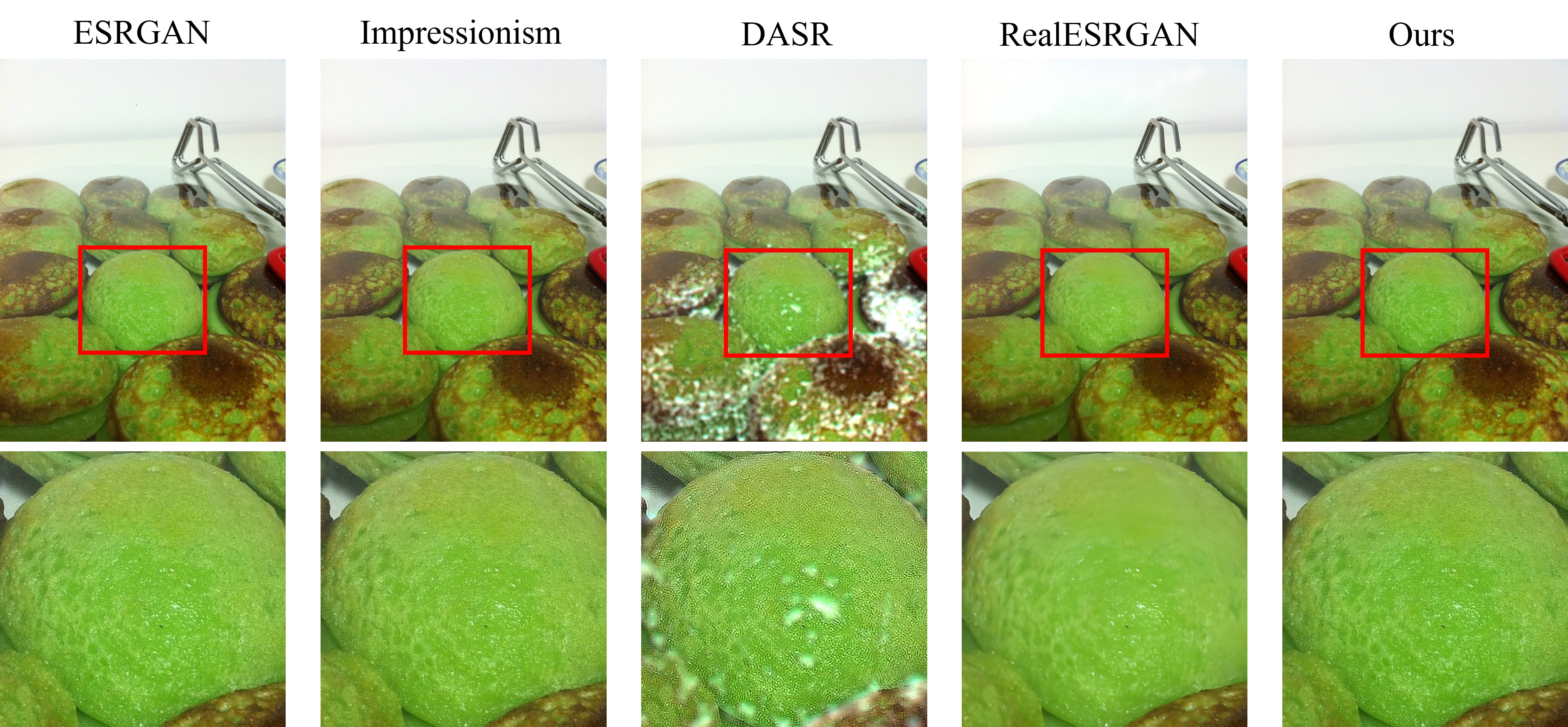}
    \caption[something short]{Super-resolution results on the RealSR-Zero dataset. We trained all methods on DF2K~\cite{Lim_2017_CVPR_Workshops} and conduct evaluation on RealSR-Zero. Particularly, RealSR-Zero is an out-of-distribution (OOD) dataset that only includes images for testing.  }
    \label{fig:realsr_zero_1}
    \vspace{-3mm}
\end{figure*}

\begin{figure*}[h]
    \centering
    \includegraphics[width=0.92\textwidth]{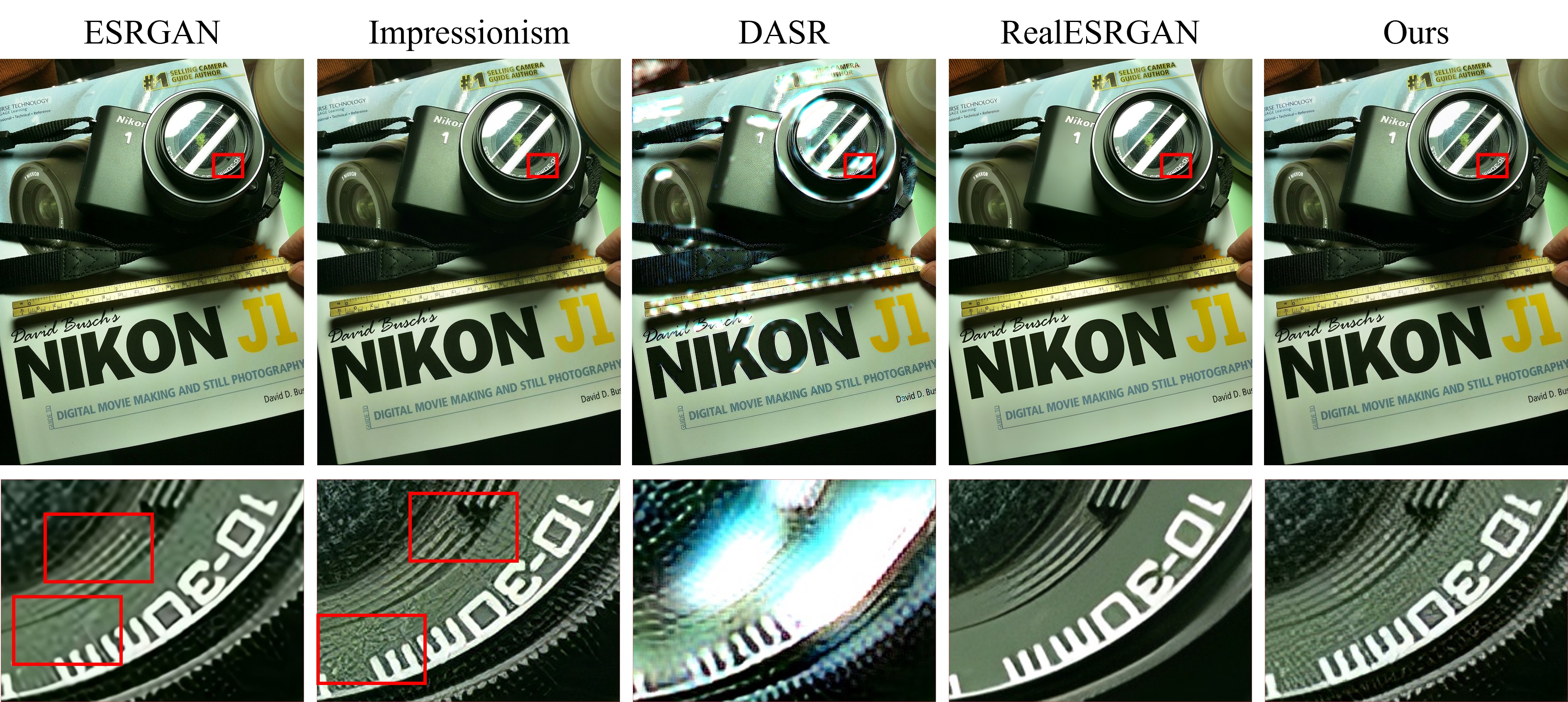}
    \caption[something short]{Super-resolution results on the RealSR-Zero dataset.}
    \label{fig:realsr_zero_2}
\end{figure*}

\begin{figure*}[t]
    \centering
    \includegraphics[width=0.99\textwidth]{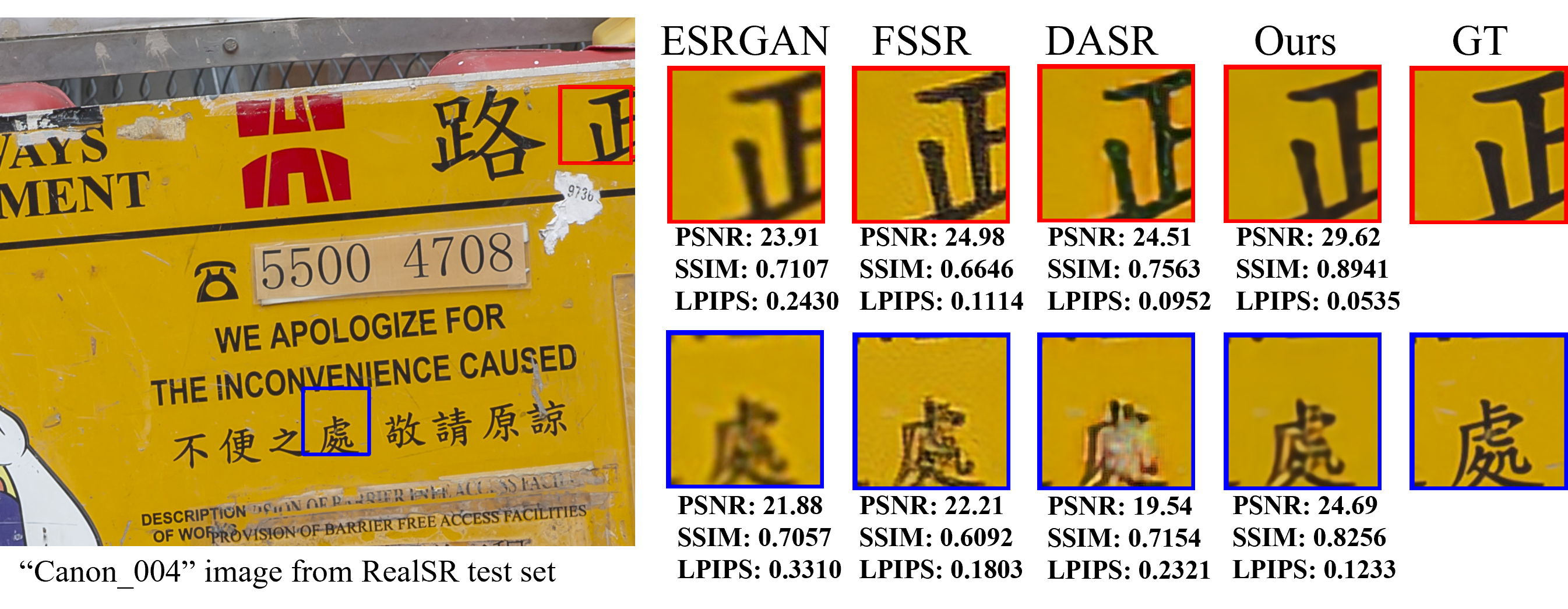}
    \caption{Super-resolution results on the RealSR~\cite{realsr} dataset.}
    \label{fig:realsr}
\end{figure*}

\begin{figure*}[h]
    \centering
    \includegraphics[width=0.99\textwidth]{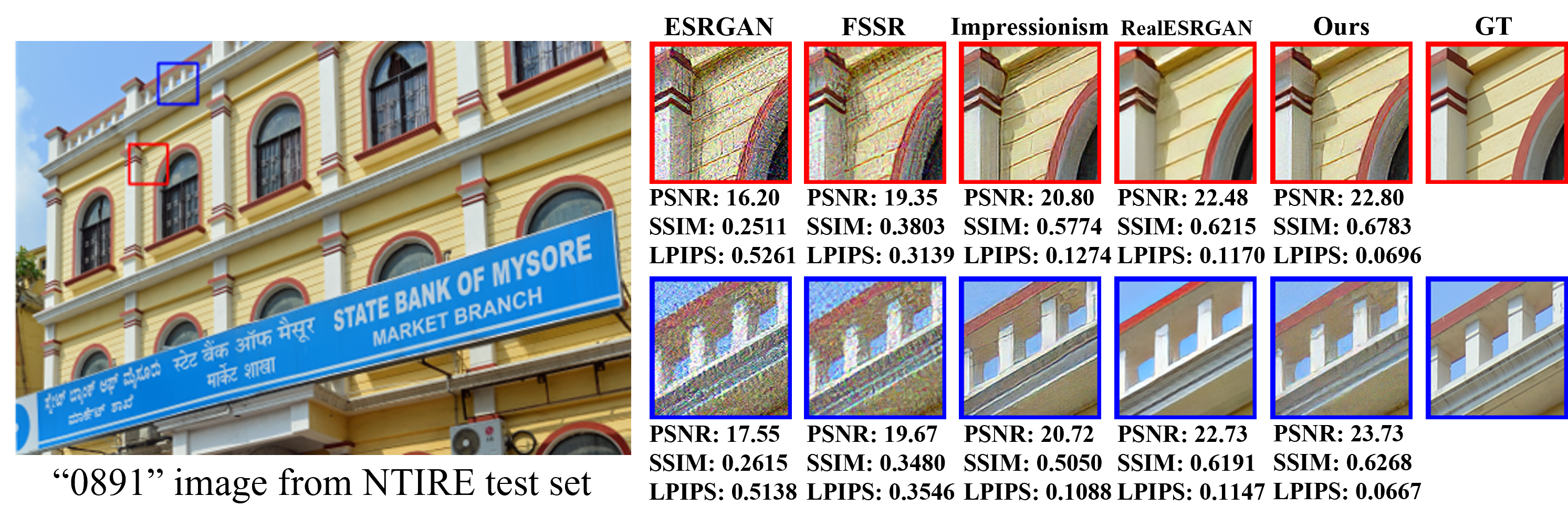} 
    \caption{Super-resolution results on the NTIRE2020 RealSR challenge data~\cite{Lugmayr2020ntire}.}
    \label{fig:ntire20}
\end{figure*}
\begin{figure*}[h]
    \centering
    \includegraphics[width=0.99\textwidth]{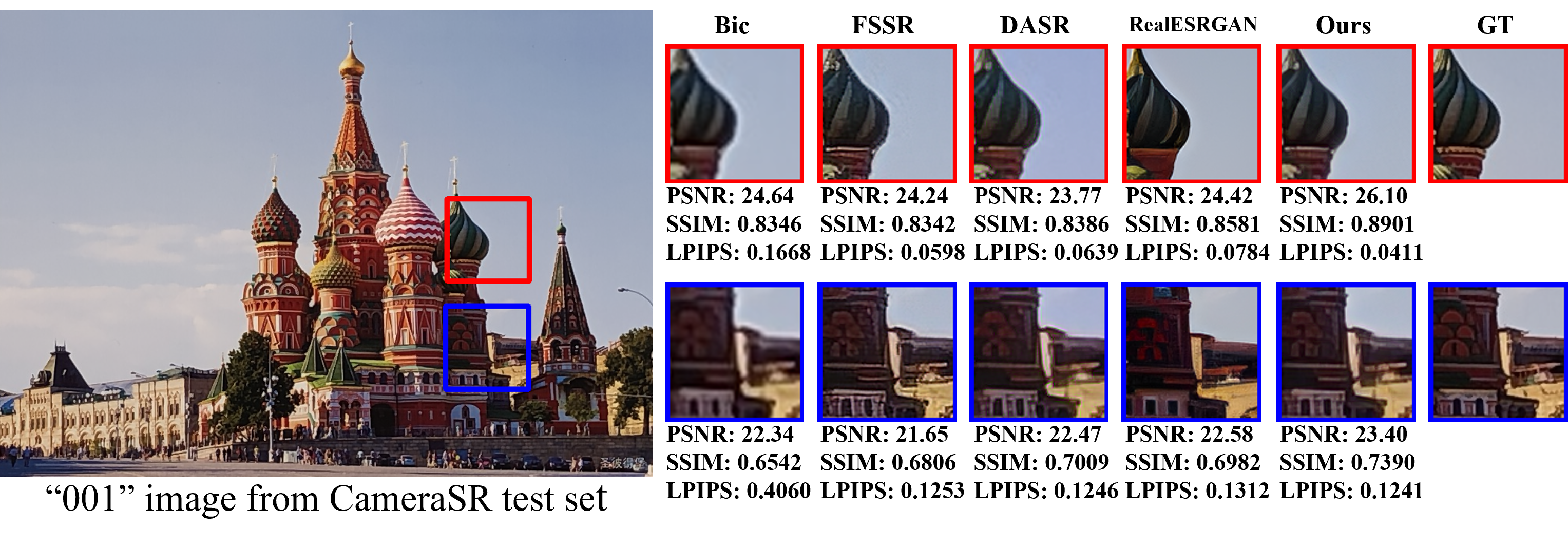}
    \caption{Super-resolution results on the CameraSR~\cite{CameraSR} dataset.}
    \label{fig:camerasr}
\end{figure*}

\section{Experiments}
\subsection{Datasets and Implementation Details}
We use following real-scene SR datasets for comprehensive comparsions to validate our model:
\begin{itemize}
  \item \textbf{\emph{RealSR-Zero}}
 consists of 45 LR images, which are shot by a iPhone4 device in different time, place and user. We collect them from internet, and the shooting period is 2011-2013 year. To modeling a challenge real-world scene, only poor-quality image are provided for evaluation. Thus, we adopt label-free quality assessment metric NIQE~\cite{mittal2012making}, to verity each method.
  \item \textbf{\emph{RealSR}}~\cite{realsr} 
 consists of 595 LR-HR image pairs. We use 200 RealSR image pairs and 800 clean images from DIV2K for training. Then, 50 LR-HR pairs, which collected by Canon camera, are used for testing. We adopt $\times$4 scale to evaluate our model.
  \item \textbf{\emph{NTIRE2020 Challenge}}~\cite{Lugmayr2020ntire} contains 3550 images, which downscaled with unknown noise to simulate inherent optical sensor noise. In our experiment, we use 3450 images, which consists of 2650 source domain images from Flickr2K and 800 target domain images from DIV2K, for training. The testing data contains 100 images from the DIV2K validation set with the same degradation operation as the source domain images. We adopt the $\times$4 scale to evaluate our model.
   \item \textbf{\emph{CameraSR}~\cite{CameraSR}} contains 200 LR-HR pairs collected by mobile phone and Nikon camera. In this work, we used 80 real-scene photos, which are shot by iPhone (e.g., No.021-100) and 800 clean images, which are fetched from DIV2K for training. And the rest 20 of LR-HR image pairs (No.001-020) are used for evaluation.
\end{itemize}

Our experiments are implemented by Pytorch 1.4.0 with 4 NVIDIA Tesla V100 GPUs. We use Adam~\cite{adam} as optimizer, where $\beta_1 =  0.9$ and $\beta_2 = 0.999$. The batchsize and total iterations are set to 16 and $10^6$, respectively. The initial learning rate is $1 \times 10^{-4}$ and decayed by $2 \times$ at every $2 \times 10^5$ iterations. We use flip and random rotation with angles of $90^{\circ}$, $180^{\circ}$ and $270^{\circ}$ for data augmentation. In evaluation protocols, we adopted PSNR, SSIM, LPIPS~\cite{lpips} and NIQE~\cite{mittal2012making} to verify our model. Also, we evaluate the inference speed on an NVIDIA Tesla V100 GPU.

\subsection{Qualitative and Quantitative Comparison} 
\emph{\textbf{RealSR-Zero.}} To perform a comparison on RealSR-Zero, we use label-free measure index NIQE and mean opinion score (MOS) for evaluation. In Tab.~\ref{tab:realsr_zero_niqe}, Cria-CL outperforms Real-ESRGAN with 0.1666 over the NIQE index, which verify that our criteria comparative algorithm help to generates richer details with high-fidelity. We also conduct human perception study by recruiting 20 volunteers for subjective assesment on RealSR-Zero. More specific, 50 pairs of patch generated by each method were shown to volunteers for a side-by-side comparison. Its note that Cria-CL wins highest preference with a 6.25\% better MOS score than Real-ESRGAN. As shown in Fig.~\ref{fig:realsr_zero_2}, the proposed model is able to avoid over-smooth and produce realistic texture. For instance, compared with Real-ESRGAN, our algorithm restores realistic texture on the green stone as well as maintains sharp edge on Fig.~\ref{fig:realsr_zero_1}, which reveals that the spatial-view projection a appropriate view for feature disentanglement in criteria comparative learning. 

\emph{\textbf{RealSR.}} As depicted in Tab.~\ref{tab:realsr}, we present a quantitative comparison. Compared with Real-ESRGAN, Cria-CL achieves a 1.38 dB gain. Our method obtains a 0.0296 LPIPS improvement over Real-ESRGAN. Compared with ADL, Cria-CL shows a 0.92 dB, which is clear improvement on RealSR task. Moreover, our algorithm restore a clear text on the second row of Fig.~\ref{fig:realsr}, which address that the criteria comparative algorithm learns richer feature for image restoration. Essentially, Real-ESRGAN and ADL are the newest state-of-the-art works, which are published in top-tier conferences and journals. This indicates that the effectiveness of Cria-CL and the contrastive relationship among criteria is worth to be fully addressed.

\emph{\textbf{NTIRE2020 and CameraSR.}} As depicted in Tab.~\ref{tab:ntire20}, compared with USR-DA~\cite{wang2021unsupervised}, Cria-CL achieves a significant improvement with 0.81 dB PSNR and 0.0323 LPIPS gain on NTIRE2020 challenge data. Compared with Real-ESRGAN, our model shows a improvement with 1.3 dB and 0.0324 LPIPS. As depicted in Tab.~\ref{tab:camerasr}, our model outperform Real-ESRGAN with 0.963 dB and 0.002 LPIPS. As USR-DA and Real-ESRGAN are recently proposed RealSR frameworks and exhibited a high-fidelity image restoration in RealSR task. Our model still achieves a significant improvement over them, which fully address the effectiveness of the proposed criteria comparative algorithm. Apart from that, Cria-CL still achieves a good visual effect in CameraSR dataset. As show in the top row of Fig.~\ref{fig:camerasr}, other methods restore blurry texture and edges in the building roof. By contrast, our model obtains smooth texture, clear boundary and fewer artifacts, which indeed justify the effectiveness of the criteria comparative algorithm and spatial-view projection. 

\begin{figure*}[h]
    \centering
    \includegraphics[width=0.9\textwidth]{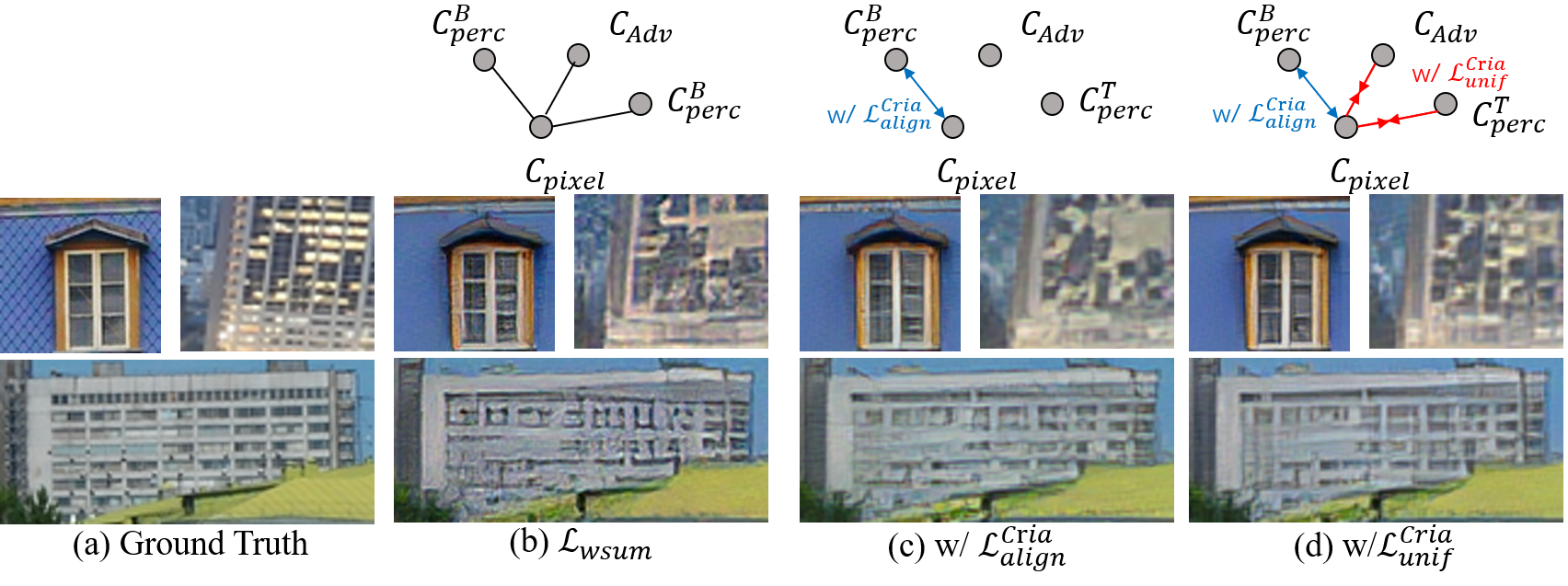}
    \caption{Visual effect of Cria-CL framework. We verify the effects of Cria-CL by presenting the visual results of each component. }
    \label{fig:abla_contrast}
\end{figure*}

%Tab.~\ref{tab:ntire20}, Tab.~\ref{tab:realsr} and Tab.~\ref{tab:camerasr} compare Cria-CL with state-of-the-art methods. In Tab.~\ref{tab:realsr}, the proposed method surpasses ADL~\cite{sonpami} with 0.92 dB PSNR. Compared with USR-DA~\cite{wang2021unsupervised}, Cria-CL achieves a significant improvement with 0.81 dB PSNR and 0.0323 LPIPS gain on NTIRE2020 challenge data. Essentially, USR-DA and ADL are the newest state-of-the-art works, which are published in top-tier conferences and journals. This indicates that the effectiveness of Cria-CL and the contrastive relationship among criteria is worth to be fully addressed.

\begin{table}[t]
\centering
\resizebox{.94\linewidth}{!}{
\begin{tabular}{lccccc}
\toprule
\multicolumn{1}{c}{Method}  & ESRGAN &Impressionism & DASR & Real-ESRGAN  & Ours         \\ \midrule \midrule
\multicolumn{1}{c}{NIQE} & 6.066 & 4.961 & 5.838 & \color{blue}{4.575} &\color{red}{4.409} \\ \midrule
\multicolumn{1}{c}{MOS} & 4.275 & 3.455 & 4.470 & \color{blue}{3.245} &\color{red}{3.050} \\ \midrule
\end{tabular}}
\caption{NIQE result for RealSR-Zero.}
\label{tab:realsr_zero_niqe}
\end{table}

\begin{table}[t]
\centering
\resizebox{.85\linewidth}{!}{
\small
\begin{tabular}{lccc}
\toprule
Methods     & \multicolumn{1}{l}{PSNR\color{red}{$\uparrow$}}            & \multicolumn{1}{l}{SSIM\color{red}{$\uparrow$}}   & \multicolumn{1}{l}{LPIPS\color{red}{$\downarrow$}} \\ \midrule \midrule
ZSSR~\cite{zeroshot}        & 26.01 & 0.7482 & 0.3861 \\
ESRGAN~\cite{wang2018esrgan} & 25.96          & 0.7468 & 0.4154 \\
CinCGAN~\cite{CinCGAN}     & 25.09                              & 0.7459           & 0.4052                     \\
FSSR~\cite{fritsche2019frequency}        & 25.99                              & 0.7388                     & 0.2653                     \\
DASR~\cite{wei2020unsupervised}        & 26.23                              & \color{blue}{0.7660}                     & \color{blue}{0.2517}           \\
Real-ESRGAN~\cite{wang2021realesrgan}      & 26.44                    & 0.7492                   & 0.2606     \\ 
ADL~\cite{sonpami}        & \color{blue}{26.90}                             & -                    & -                   \\ \midrule \midrule
Ours       & {\color{red}{27.82}}          & {\color{red}{0.8123}} & {\color{red}{0.2311}} \\ \bottomrule
\end{tabular}}
\caption{Quantitative results on RealSR dataset.}
\label{tab:realsr}
\end{table}

\begin{table}[h]
\centering
\resizebox{.85\linewidth}{!}{
\small
\begin{tabular}{lccc}
\toprule                                   
Methods     & \multicolumn{1}{l}{PSNR {\color{red}$\uparrow$}} & \multicolumn{1}{l}{SSIM\color{red}{$\uparrow$}} & \multicolumn{1}{l}{LPIPS\color{red}{$\downarrow$}} \\ \midrule \midrule
EDSR~\cite{edsr}        & 25.31           & 0.6383                   & 0.5784                    \\
ESRGAN~\cite{wang2018esrgan} & 19.06                    & 0.2423                   & 0.6442                    \\
ZSSR~\cite{zeroshot}        & 25.13                    & 0.6268                   & 0.6160                     \\
KernelGAN~\cite{kernelgan}      & 18.46                    & 0.3826                   & 0.7307                    \\
Impressionism~\cite{Ji_2020_CVPR_Workshops}      & 24.82                    &0.6619                   & \color{blue}{0.2270}     \\        
Real-ESRGAN~\cite{wang2021realesrgan}      & 24.91                    &0.6982                   & 0.2468     \\ 
USR-DA~\cite{wang2021unsupervised}      & \color{blue}{25.40}                     &\color{blue}{0.7075}                   & 0.2524                     
\\ \midrule \midrule
Ours        & {\color{red}{26.21}}                   & {\color{red}{0.7122}}          & {\color{red}{0.2201}}            \\ \bottomrule
\end{tabular}}
\caption{Quantitative results for NTIRE2020 Challenge on Real-world image SR track.}
\label{tab:ntire20}
\end{table}

\begin{table}[h]
\centering
\resizebox{.85\linewidth}{!}{
\small
\begin{tabular}{l|ccc}
\toprule
Methods     & \multicolumn{1}{l}{PSNR\color{red}{$\uparrow$}}            & \multicolumn{1}{l}{SSIM\color{red}{$\uparrow$}}   & \multicolumn{1}{l}{LPIPS\color{red}{$\downarrow$}} \\ \midrule \midrule
FSSR~\cite{fritsche2019frequency}        & 23.781                              & 0.7566                     & 0.180    \\
{Impressionism}~\cite{Ji_2020_CVPR_Workshops}        & {25.142}              & {\color{blue}{0.8097}}             & 0.139                    \\
DASR~\cite{wei2020unsupervised}       & \color{blue}{25.235}                              & 0.8065                     & 0.141                   \\
Real-ESRGAN~\cite{wang2021realesrgan}      & 25.175                    &0.8023                   & \color{blue}{0.136}     \\ \midrule \midrule
%Ours        & \underline{26.242}                              & \underline{0.8224}                     & \underline{0.136}                     \\
Ours       & \color{red}{26.138}          & \color{red}{0.8172} & \color{red}{0.134} \\ \bottomrule
\end{tabular}}
\caption{Quantitative results on CameraSR dataset.}
\label{tab:camerasr}
\end{table}

\subsection{Ablations} We conduct extensive ablations of our Cria-CL framework on NTIRE 2020 challenge data to verify the effectiveness of each component.

\emph{\textbf{Criteria Comparative Algorithm.}} In Tab.~\ref{table:ablation}, a plain model achieves 0.3 dB gain by using $\mathcal{L}^{Cria}_{align}$ which verifies the effectiveness of alignment loss between positive losses. For a fair comparison, we apply $\mathcal{L}^{Cria}_{unif}$ on the `Baseline' model, which achieves a significant promotion with a 0.94 dB gain. We conduct this quantitative analysis on NTIRE2020 dataset, which shows that each proposed component is necessary for our model.

\emph{\textbf{Visual Effects.}} We also illustrate the visual effect of each proposed component in Fig.~\ref{fig:abla_contrast}. Obviously, the enhanced images by $\mathcal{L}^{Cria}_{unif}$ restore more correct details with fewer artifacts, indicating that the uniformity constraint can significantly improve the visual qualities under the multi-task paradigm.

\begin{table}[t]
\centering
\resizebox{.95\linewidth}{!}{
\small
\begin{tabular}{lccc}
\toprule
Methods &PSNR\color{red}{$\uparrow$} &SSIM\color{red}{$\uparrow$} &LPIPS\color{red}{$\downarrow$}\\
\midrule \midrule
Plain Model &24.82 &0.6619 & 0.2270\\  \midrule 
w/ $\mathcal{L}^{Pixel}$ + $\mathcal{L}^{perc}$ +  $\mathcal{L}^{adv}$ &{25.05}  &{0.6683} &{0.2269} \\
w/ $\mathcal{L}^{pixel}$ + $\mathcal{L}^{perc}$ + $\mathcal{L}^{adv}$ + $\mathcal{L}^{Cria}_{align}$ & 25.12 &0.6720 &0.2263\\
w/ $\mathcal{L}^{pixel}$ + $\mathcal{L}^{perc}$ + $\mathcal{L}^{adv}$ + $\mathcal{L}^{Cria}_{unif}$ &25.76& 0.6977 & \color{blue}{0.2227} \\
w/ $\mathcal{L}^{pixel}$ + $\mathcal{L}^{perc}$ + $\mathcal{L}^{adv}$ + $\mathcal{L}^{Cria}_{align}+\mathcal{L}^{Cria}_{unif}$ &{25.92} &{0.6983} &{0.2230} \\
w/ Spatial-view Projection +  $\mathcal{L}^{pixel}$ + $\mathcal{L}^{perc}$ + $\mathcal{L}^{adv}$ & \color{blue}{26.04} &\color{blue}{0.7042} & 0.2241\\ 
\midrule \midrule
Ours full  & {\color{red}{26.21}}          & {\color{red}{0.7122}} & {\color{red}{0.2201}} \\
\bottomrule
\end{tabular}}
% \vspace{-0.5em}
\caption{Ablation study of Cria-CL. Each proposed component is analysed independently to show its effect. }
\label{table:ablation}
% \vspace{-1em}
\end{table}

\begin{table}[t]
\centering
%\large
\begin{tabular}{lcc}
\toprule
Methods &PSNR\color{red}{$\uparrow$}  &LPIPS\color{red}{$\downarrow$}\\
\midrule \midrule
$\mathbb{C}_{a} = \mathbb{C}_{adv}$  &17.97  &0.6322 \\
$\mathbb{C}_{a} = \mathbb{C}_{per}$  &24.27  &0.3924 \\
$\mathbb{C}_{a} = \mathbb{C}_{pix}$ &26.21  &0.2201 \\
\bottomrule
\end{tabular}
% \vspace{-0.5em}
\caption{Quantitative comparison for different anchor criterion on NTIRE2020 dataset. }
\label{table:ablation_anchor}
% \vspace{-1em}
\end{table}

\begin{table}[t]
\centering
\resizebox{.95\linewidth}{!}{
\begin{tabular}{lcccc}
\toprule
\multicolumn{1}{c}{Method}  & ESRGAN & DASR & Real-ESRGAN  & Ours         \\ \midrule \midrule
\multicolumn{1}{c}{Time(frame/s)} & 0.7971 & 0.7465 & 0.7652 & 0.7217 \\ \midrule \midrule
\multicolumn{1}{c}{Parameter} & 16.7M &16.7M  & 16.7M & 14.8M \\ \bottomrule
\end{tabular}}
\caption{Efficiency analysis on 300$\times$200 image of RealSR testset with $4 \times$ factor. As DASR and Real-ESRGAN adopt ESRGAN as backbone, they have same network parameters.}
\label{tab:efficiency}
\end{table}

\emph{\textbf{Spatial-view Projection.}} We show the effect of spatial-view projection in Tab.~\ref{table:ablation}. With the spatial-view mechanism, our model obtains a 1.19 dB improvement. Without spatial-view projection, $\mathcal{L}^{Cria}_{unif}$ and  $\mathcal{L}^{Cria}_{align}$ exhibit limited performance improvement. This shows that in the Cria-CL framework, spatial-view projection is required for a good view of feature disentanglement among multi-criteria training conditions. 

%\begin{figure*}
%    \centering
%    \includegraphics[width=0.95\textwidth]{fig/vc3.png} 
%    \caption[something short]{Super-resolution results on the NTIRE2020 RealSR challenge data~\cite{Lugmayr2020ntire}.}
%    \label{fig:vc3}
%\end{figure*}

% Caption should have no citation

\subsection{Efficiency.} We conduct the efficiency analysis toward the state-of-the-art methods on the RealSR dataset with their official implementation and equal hardware environment. As shown in Tab.~\ref{tab:efficiency}, Cria-CL achieves competitive efficiency with attractive performance. Specifically, Cria-CL obtains a faster running efficiency with a 1.83 dB improvement over FSSR. Compared with ESRGAN, Cria-CL obtains promising running efficiency by reducing 10\% inference cost. Compared with DASR, the proposed Cria-CL demonstrates superior efficiency and achieves significant improvements by 1.57 dB. This shows that the Cria-CL learns effective feature representations for RealSR with fewer parameters.

\section{Discussion}
As the proposed Cria-CL shows promising results on the RealSR task, a few open problems still need to be further explored. Cria-CL sets the pixel loss as the anchor and achieves attractive performance. Nevertheless, when Cria-CL uses adversarial loss as the anchor for contrastive multi-task learning, the performance becomes worse. This suggests the positive counterpart toward the adversarial criterion required for further investigation.

Except for the RealSR task, Cria-CL has the potential to be applied to other real-world image processing tasks, such as de-raining, image enhancement, and de-hazing. We hope Cria-CL will bring diverse insight to the image processing tasks that include contrastive learning.
% \appendix

\bibliographystyle{IEEEtran}
\bibliography{egbib}

\end{document}